\documentclass[sigconf]{acmart}
\AtBeginDocument{%
	\providecommand\BibTeX{{%
			\normalfont B\kern-0.5em{\scshape i\kern-0.25em b}\kern-0.8em\TeX}}}

\setcopyright{none}
\setcopyright{acmcopyright}

\copyrightyear{2024} 
\acmYear{2024} 
\setcopyright{acmlicensed}\acmConference[KDD '24]{Proceedings of the 30th ACM SIGKDD Conference on Knowledge Discovery and Data Mining}{August 25--29, 2024}{Barcelona, Spain}
\acmBooktitle{Proceedings of the 30th ACM SIGKDD Conference on Knowledge Discovery and Data Mining (KDD '24), August 25--29, 2024, Barcelona, Spain}
\acmDOI{10.1145/3637528.3671700}
\acmISBN{979-8-4007-0490-1/24/08}

\newcommand{\specialcell}[2][c]{%
	\begin{tabular}[#1]{@{}c@{}}#2\end{tabular}}

\newcommand{\bb}{\hspace{-1mm} $\bullet$}

\usepackage[ruled]{algorithm2e}
\usepackage{multirow}
\usepackage{arydshln}
\usepackage{subcaption}
\usepackage{booktabs}
\usepackage{amsmath}

\usepackage{pifont}

\usepackage{colortbl}

\definecolor{lightgrey}{rgb}{0.9,0.9,0.9}

\newtheorem{theorem}{Theorem}
\newtheorem{lemma}{Lemma}

\newcommand{\ours}{GeoMix\xspace}
\newcommand{\cmark}{\ding{51}}%
\newcommand{\xmark}{\ding{55}}%

\begin{document}
\title{
GeoMix: Towards Geometry-Aware Data Augmentation
}
	
	
	\author{Wentao Zhao}
        \email{permanent@sjtu.edu.cn}
        \affiliation{%
		\institution{Shanghai Jiao Tong University}
            \department{Department of Computer Science and Engineering\\MoE Key Lab of Artificial Intelligence}
		\city{Shanghai}
		\country{China}
        }  

        \author{Qitian Wu}
        \email{echo740@sjtu.edu.cn}
	\affiliation{%
		\institution{Shanghai Jiao Tong University}
            \department{Department of Computer Science and Engineering\\MoE Key Lab of Artificial Intelligence}
		\city{Shanghai}
		\country{China}
	}

        \author{Chenxiao Yang}
        \email{chr26195@sjtu.edu.cn}
	\affiliation{%
		\institution{Shanghai Jiao Tong University}
            \department{Department of Computer Science and Engineering\\MoE Key Lab of Artificial Intelligence}
		\city{Shanghai}
		\country{China}
	}

        \author{Junchi Yan}
        \email{yanjunchi@sjtu.edu.cn}
        \authornote{Corresponding author.}
	\affiliation{%
		\institution{Shanghai Jiao Tong University}
            \department{Department of Computer Science and Engineering\\MoE Key Lab of Artificial Intelligence}
		\city{Shanghai}
		\country{China}
	}
        
	\renewcommand{\shortauthors}{Wentao Zhao, Qitian Wu, Chenxiao Yang, \& Junchi Yan}
	
	\begin{abstract}
		Mixup has shown considerable success in mitigating the challenges posed by limited labeled data in image classification. By synthesizing samples through the interpolation of features and labels, Mixup effectively addresses the issue of data scarcity. However, it has rarely been explored in graph learning tasks due to the irregularity and connectivity of graph data. Specifically, in node classification tasks, Mixup presents a challenge in creating connections for synthetic data. In this paper, we propose Geometric Mixup (GeoMix), a simple and interpretable Mixup approach leveraging in-place graph editing. It effectively utilizes geometry information to interpolate features and labels with those from the nearby neighborhood, generating synthetic nodes and establishing connections for them. We conduct theoretical analysis to elucidate the rationale behind employing geometry information for node Mixup, emphasizing the significance of locality enhancement—a critical aspect of our method's design. Extensive experiments demonstrate that our lightweight Geometric Mixup achieves state-of-the-art results on a wide variety of standard datasets with limited labeled data. Furthermore, it significantly improves the generalization capability of underlying GNNs across various challenging out-of-distribution generalization tasks. Our code is available at \url{https://github.com/WtaoZhao/geomix}.
	\end{abstract}


	\maketitle
	
	\section{Introduction}
	Graph Neural Networks (GNNs) \cite{gcn, gat, sgc, appnp} have become the de facto method for modeling the increasingly popular graph-structured data. However, in real world, labeling data is expensive and many datasets have very few labeled examples. This scarcity of labeled data can lead to severe over-fitting issues and weaken the generalization performance of GNNs, especially when the test data comes from a distribution that differs from the training data, which is a common scenario in many real-world datasets.
	
	Motivated by the above issues, we set out to design Mixup for graph learning, a technique that has demonstrated substantial success in mitigating challenges caused by limited data and enhancing model performance \cite{ZhangCDL18}.  At its core, Mixup trains neural networks on convex combinations of pairs of examples and their corresponding labels.  This approach broadens the distribution of training data and regularizes neural networks, serving as the key factors behind its capacity to reduce over-fitting, facilitate the learning of more discriminative representations, and enhance model generalization \cite{zhang2020does}. These attributes are pivotal when handling datasets with limited labeled data or where the training data only encompass a subset of the diverse data distributions, which might not fully represent the distributions of the testing data.
	
	Though prevailingly used in other fields, Mixup has rarely been explored in graph learning tasks, due to the connectivity in graph. In node classification task, questions have arisen about how to effectively connect synthetic nodes. Current general Mixup strategies \cite{verma2021graphmix, wang2021mixup} often attempt to circumvent the explicit connection of synthetic nodes by performing Mixup between layers of neural networks, potentially limiting their adaptability. Other Mixup methods aimed at addressing class-imbalance problems incorporate complex edge prediction modules. Unfortunately, this sacrifices Mixup's inherent lightweight nature and may diminish the generalization power due to increased model complexity.
	
	To address these challenges, this paper proposes a simple and geometry-aware Mixup approach that in-place modifies raw data and explicitly establishes connections for synthetic nodes, enhancing interpretability. Theoretical analysis on the mixed features and labels reveals: (1) the interpolation effects of this geometry-aware Mixup; (2) its rationale for utilizing geometry information; (3) scenarios where this fundamental strategy may succeed or fail. 
	
	Building upon the theoretical analysis and recognizing potential failure cases, we further refine our approach and present Geometric Mixup.
	It not only considers geometry details but also enhances locality information, enabling it to adapt to both homophilic graphs (where adjacent nodes are likely to have similar labels and features) and heterophilic graphs (adjacent nodes tend to have dissimilar labels). 
	Moreover, we elucidate the connection between Geometric Mixup and graph structure learning to provide more insight and better interpretability for our design. 
	
	Extensive experiments across twelve datasets demonstrate that: (1) Geometric Mixup achieves state-of-the-art results on both homophilic and heterophilic graphs with limited labeled data; (2) it significantly improves the generalization ability of underlying GNNs in various out-of-distribution generalization tasks with limited distributions of training data; (3) it assists underlying GNNs in learning more discriminative representations, improving prediction performance. 
 
 \textbf{The major contributions of our work are:}
	
	1) We propose a simple and interpretable Mixup strategy leveraging in-place graph editing, which is a novel perspective.
	
	2) Our approach effectively utilizes graph geometry while enhancing locality information to accommodate to both homophilic and heterophilic graphs.
	
	3) Theoretical analysis provides insights into leveraging geometry information for Mixup and underlines the significance of enhancing locality information. 
	
	4) Extensive experiments substantiate that Geometric Mixup effectively improves the performance and generalization of underlying GNNs in challenging tasks with limited training data. 
	
	To distinguish our approach from existing ones, we compare Geometric Mixup with other node Mixup methods in Table \ref{tab:compare}.
	
	\begin{table*}[tb!]
	\caption{Comparison of GeoMix (short for Geometric Mixup) with other node Mixup.}
	\label{tab:compare}
	\centering
        \begin{tabular}{lccccc}
        \toprule
        Method & \specialcell{Use geometry\\information} & \specialcell{Explicitly connect\\synthetic nodes} & \specialcell{Support modification\\to raw data} & \specialcell{Introduce time-consuming\\edge prediction module} & \specialcell{Support OOD\\Generalization} \\
        \midrule
        GraphMix \cite{verma2021graphmix} & \xmark & \xmark & \xmark & \xmark & \cmark \\
        Mixup \cite{wang2021mixup} & \xmark & \xmark & \xmark & \xmark & \cmark \\
        GraphMixup \cite{wu2022graphmixup} & \xmark & \cmark & \cmark & \cmark & \xmark \\
        GeoMix (ours) & \cmark & \cmark & \cmark & \xmark & \cmark \\
        \bottomrule
        \end{tabular}
        \end{table*}
	
	\section{Preliminaries}
	\subsection{Semi-supervised Node Classification}
	Let  $G=(\mathcal{V}, \mathcal{E})$ denotes a graph with node set $ \mathcal{V} $ and edge set $ \mathcal{E} $. Each node $ v \in \mathcal{V} $ is associated with a feature vector $ \mathbf{x}_v $ and a  label $ y_v $, represented in one-hot form as $ \mathbf{y}_v $. Denote node feature matrix as $ \mathbf{X}=\{\mathbf{x}_i\}_{i=1}^{|\mathcal{V}|} $, adjacency matrix as $ \mathbf{A} $.
	The goal of node classification task is to train a classifier $ f(\cdot) $ that can accurately predict node labels based on $ \mathbf{X} $ and $ \mathbf{A} $.
	In semi-supervised setting, the classifier has access to the complete feature matrix $ \mathbf{X} $ and adjacency matrix $ \mathbf{A} $, but is restricted to having labels for only a subset of nodes, constituting the labeled node set $ \mathcal{V}_l $ (we denote the unlabeled node set as $ \mathcal{V}_u $). 
	Therefore, the standard loss function for semi-supervised node classification is
	\begin{align}
		\sum_{v \in \mathcal{V}_l } \ell (f(\mathbf{A}, \mathbf{X})_v , \mathbf{y}_v ),
	\end{align}
	where $ \ell $ is usually cross-entropy loss. $ f(\mathbf{A}, \mathbf{X})_v  $ is the prediction for node $ v $.
	
	\subsection{Message Passing Neural Networks}
	Message passing neural networks propagate information along edges to learn node representations, which can be expressed as: 
	\begin{align}
		\mathbf{h}_v^{(k+1)}=\mathsf{AGGR}\left( \mathbf{h}_v^{(k)},  \{\mathbf{h}_u^{(k)}: u \in {\mathcal{N}}(v) \}    \right) ,
	\end{align}
	where $\mathcal{N}(v) $ is the neighborhood of $ v $.
	$ \mathsf{AGGR} $ function aggregates information from neighboring nodes and combines the results with the current state of the central node to update its representation.  
	
	\subsection{Mixup}
	Mixup is first proposed for image classification \citep{ZhangCDL18}. It linearly mixes both features and labels of samples, which can be written as
	\begin{align}
		\bar{\mathbf{x}}&=\lambda  \mathbf{x}_i + (1-\lambda) \mathbf{x}_j, \label{eq:image_x_mixup} \\
		\bar{\mathbf{y}}&=\lambda  \mathbf{y}_i + (1-\lambda) \mathbf{y}_j, \label{eq:image_y_mixup}
	\end{align}
	where $ i, j $ is a random pair of samples, $ \lambda \in [0, 1] $. However, adapting Mixup to node classification is nontrivial due to the challenge in defining neighborhood for synthetic nodes. 
	
	\section{Methods}
	\subsection{A Basic Geometry-Aware Mixup} \label{sec:gcn_mixup}
	Though it may seem natural to apply Eq. \eqref{eq:image_x_mixup} and \eqref{eq:image_y_mixup} to node features and node labels to create synthetic nodes, how to establish connections for synthetic nodes remains an unsolved problem. Inspired by message passing which iteratively updates node features by combining information from neighboring nodes, we propose an in-place-editing-based Mixup, where a node's feature/label is adjusted using a convex combination of features/labels from its immediate neighborhood. It explicitly connects synthetic nodes without necessitating a complex edge prediction module and effectively leverages prior knowledge from the given graph.

	However, one challenge of semi-supervised learning lies in the scarcity of ground truth labels, leading to incomplete or inaccessible label information for neighborhoods of certain nodes. To this end, we first employ the training model $ f(\cdot) $ to predict the pseudo label for each unlabeled node. For the convenience of subsequent derivation, denote $ \hat{\mathbf{y}}_v $ as:
	\begin{align}
		\hat{\mathbf{y}}_v=\begin{cases}
			\mathbf{y}_v& \text{if $ v\in \mathcal{V}_l $}\\
			f(\mathbf{A},\mathbf{X})_v&\text{otherwise}		
		\end{cases}
	\end{align}
	The most basic approach to leverage geometry information in Mixup involves updating a node's feature and label to be the average of its neighbors' features and labels. Consequently, the mixing operation for an arbitrary node $ v $ can be expressed as:
	\begin{align}
		\mathbf{h}^{(t+1)}_v&= \sum_{u \in {\mathcal{N}}(v) } e_{vu} \mathbf{h}^{(t)}_u, \label{eq:geomix_basic_x} \\
		\bar{\mathbf{y}}^{(t+1)}_v&= \sum_{u \in {\mathcal{N}}(v) } e_{vu} \bar{\mathbf{y}}^{(t)}_u, \label{eq:geomix_basic_y}
	\end{align}
	where $ \mathbf{h}^{(t)}_v $ and $ \bar{\mathbf{y}}^{(t)}_v $ are the feature and
	label of node $ v $ after $ t $-th operation. $ \mathbf{h}^{(0)}_v=\mathbf{x}_v $ is the input node feature. $\bar{\mathbf{y}}^{(0)}_v=\hat{\mathbf{y}}_v $. $ e_{uv} $ is the edge weight given by common normalized adjacency matrix like $ \mathbf{D}^{-1} \mathbf{A} $ and $ \mathbf{D}^{-1/2} \mathbf{A}\mathbf{D}^{-1/2} $ where $ \mathbf{D} $ is the degree matrix. 
	We conduct theoretical analysis on  Eq. \eqref{eq:geomix_basic_x} and \eqref{eq:geomix_basic_y} to demonstrate its interpolation effects and offer insights into the circumstances under which this fundamental form of Mixup may succeed or fail.
	
	\textbf{Assumptions on Graphs.}
	To ease the analysis, we pose the following assumptions on graphs. Denote  the number of classes as $ C $. Assume that for any node $ i $: (1) Its feature $ \mathbf{x}_i $ is sampled from feature distribution $ \mathcal{D}_{y_i} $ associated with its label, with $ \boldsymbol{\mu}(y_i) $ denoting its mean; (2) dimensions of $ \mathbf{x}_i $ are independent to each other; (3) the feature values in $ \mathbf{x}_i $ are bounded by a positive scalar $ B $, i.e., $ \max_k | \mathbf{x}_i[k]| \leq B $; (4) due to lack of ground truth labels and errors in pseudo label prediction, the expectations of $ \hat{\mathbf{y}}_i $ is 
	\begin{align}
		\mathbb{E}[\hat{\mathbf{y}}_i]=(1-\epsilon) \mathbf{e}_{y_i} + \frac{\epsilon}{C-1}\sum_{j\neq y_i} \mathbf{e}_{j},
	\end{align}
	where $ \mathbf{e}_{y_i} $ is the $ y_i $-th standard basis vector (all elements are 0 except that the $ y_i $-th element is 1), $ \epsilon \in (0,1) $ is related to the label rate and the accuracy of model used to predict pseudo-labels. When the label rate is large or the model is highly dependable, $ \epsilon $ should be close to 0; (5) Its neighbors' labels $ \{y_j : j\in \mathcal{N}(i) \} $  are conditionally independent given $ y_i $, and have the same label as node $ i $ with probability $ p $. They belong to any other class $ c \neq y_i $ with probability $ (1-p)/(C-1) $. 
	
	We use $ \mathcal{G}=\{\mathcal{V}, \mathcal{E}, \{\mathcal{D}_c, c\in C  \}, p, \epsilon \}$ to denote a graph following the above assumptions. Note that we use subscript $ c $ to indicate that distribution $ \mathcal{D}_c $ is shared by all nodes with the same label $ c $. Then we have the following theorem about mixed features:
	\begin{theorem} \label{th:mixed_feature}
		Consider a graph $  \mathcal{G}=\{\mathcal{V}, \mathcal{E}, \{\mathcal{D}_c, c\in C  \}, p, \epsilon \} $ following Assumptions (1)-(5). For any node $ i \in \mathcal{V} $, the expectation of its feature after performing one Mixup operation is
		\begin{align}
			\mathbb{E}[\mathbf{h}_i]=p \mathbf{\boldsymbol{\mu}}(y_i) + \frac{1-p}{C-1} \sum_{c \neq y_i} \mathbf{\boldsymbol{\mu}}(c), \label{eq:exp_feature}
		\end{align} 
		and for any $ t>0 $,  the probability that the distance between the observation $ \mathbf{h}_i $ and its expectation is larger than t is bounded by
		\begin{align}
			\mathbb{P}(\| \mathbf{h}_i - \mathbb{E}[\mathbf{h}_i ]  \|_2 \geq t  ) \leq 2F\exp\left( -\frac{deg(i)t^2 }{2B^2F}   \right), \label{eq:feature_distance}
		\end{align}
		where $ F $ is  the feature dimension.
	\end{theorem}
	Similarly, for mixed labels, we have
	\begin{theorem} \label{th:mixed_label}
		For any $ c\in C $ and any  $ i \in \mathcal{V} $ with $ y_i=c $, the expectation of mixed label $ \bar{\mathbf{y}}_i $ after performing one Mixup operation is
		\begin{align}
			\begin{aligned}
			\mathbb{E}[\bar{\mathbf{y}}_i]	= \left[ p(1-\epsilon)+ \frac{\epsilon(1-p)}{C-1}   \right] \mathbf{e}_c+ \left[\frac{p\epsilon+(1-p)(1-\epsilon)}{C-1} \right. \\
			\left. +\frac{\epsilon(1-p)(C-2)}{(C-1)^2} \right] \sum_{j\neq c} \mathbf{e}_j,
			\end{aligned}
		 \label{eq:exp_label}
		\end{align}
		and for any $ t>0 $,  the probability that the distance between the observation $ \bar{\mathbf{y}}_i $ and its expectation is larger than t is bounded by
		\begin{align}
			\mathbb{P}(\| \bar{\mathbf{y}}_i- \mathbb{E}[\bar{\mathbf{y}}_i]  \|_2 \geq t  ) \leq 2C\exp\left( -\frac{deg(i)t^2 }{2C}   \right) \label{eq:label_dist}.
		\end{align} 
	\end{theorem}
	The proofs of Theorem \ref{th:mixed_feature} and \ref{th:mixed_label} can be found in the appendix. The above theorems demonstrate two facts. Firstly, when $ p $ is large and $ \epsilon $ is small, the mixed feature and label of node $ i $ will stay comparatively close to its input feature and label in expectation. Secondly, the distance between the mixed feature/label of node $ i $ and its expectation is small with a high probability.
	Together, they show that the locality information is preserved in the Mixup and justify the rationale to place the updated node in its previous position. 
	Furthermore, within Eq. \eqref{eq:exp_feature} and \eqref{eq:exp_label}, we observe the desired interpolation effects achieved through Mixup.

	However, when $ p $ is small, as is in some challenging heterophilic graphs, the mixed feature/label of node $ i $ will be far from its original feature/label in expectation. Thus, the locality information is not well-preserved and it becomes dubious to place the updated node in its original position. Another problem arising from a small value of $ p $ is the decreased distinguishability in the expectations of mixed features/labels from nodes belonging to different classes. In extreme cases when $ p\to 1/C $, the expected features/labels of nodes from different classes converge to the same point, which greatly diminishes the diversity of mixed features and labels. Therefore, this basic geometry-aware Mixup may fail to perform well in some heterophilic graphs, and we will provide solutions next.
	
	\subsection{Geometric Mixup: Locality-Enhanced Mixup with Structure Awareness} \label{sec:geomix}
	\begin{figure*}[tb!]
      \centering
      \includegraphics[width=\linewidth]{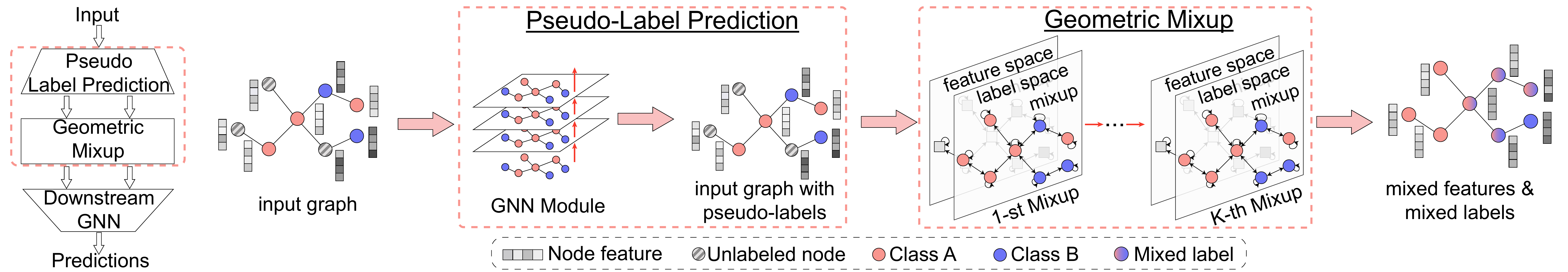}
      \caption{Illustration of the training procedure with Geometric Mixup.}
      \label{fig:geomix}
    \end{figure*}
	One feasible solution to the problems elucidated in the preceding section is to enhance the locality information by adding residual connections. For node $ v $, we establish a residual connection utilizing $ \mathbf{h}_v^{(t)} $, its mixed feature from the preceding Mixup operation. This locality fortification yields the first variant of Geometric Mixup, which is expressed in Eq. \eqref{eq:geomix_gcn_x}-\eqref{eq:geomix_gcn_y} and illustrated in Fig. \ref{fig:geomix}.
	\begin{align}
		\mathbf{h}^{(t+1)}_v&=\alpha \mathbf{h}^{(t)}_v+(1-\alpha) \sum_{u \in {\mathcal{N}}(v) } e_{vu} \mathbf{h}^{(t)}_u, \label{eq:geomix_gcn_x} \\
		\bar{\mathbf{y}}^{(t+1)}_v&=\alpha \bar{\mathbf{y}}^{(t)}_v+(1-\alpha) \sum_{u \in {\mathcal{N}}(v) } e_{vu} \bar{\mathbf{y}}^{(t)}_u. \label{eq:geomix_gcn_y}
	\end{align} 
	The residual connection $ \mathbf{h}_v^{(t)} $ encompasses information from nodes within a distance of $ t $ hops from node $ v $ and thus helps to better preserve the locality information of node $ v $.
	$ \alpha $ is a hyper-parameter controlling the effect of locality reinforcement.
	By choosing a proper value for $ \alpha $, the expectation of $ \mathbf{h}_v^{(t+1)} $ will stay relatively close to its preceding value. Consequently, the expected features/labels of nodes from different classes will not converge to the same point even when $ p $ is small and the locality information is effectively preserved. Therefore, the diversity of synthetic data will not be compromised and it is reasonable for the updated node to remain in its position.
	Note that we may repeat the Mixup operation for $ K $ times to add more comprehensive range of geometry information to the mixed features and labels. In practice, one or two consecutive Mixup achieves good performance. 
	
	A more radical and effective choice of preserving  locality information from the input graph is to utilize node $ v $'s original feature $ \mathbf{h}_v^{(0)} $ and label $ \bar{\mathbf{y}}^{(0)}_v $  to establish the residual connection.
	\begin{align}
		\mathbf{h}^{(t+1)}_v&=\alpha \mathbf{h}^{(0)}_v+(1-\alpha) \sum_{u \in {\mathcal{N}}(v)  } e_{vu} \mathbf{h}^{(t)}_u \label{eq:geomix_app_x},  \\ 
		\bar{\mathbf{y}}^{(t+1)}_v&=\alpha \bar{\mathbf{y}}^{(0)}_v+(1-\alpha) \sum_{u \in {\mathcal{N}}(v) } e_{vu} \bar{\mathbf{y}}^{(t)}_u. \label{eq:geomix_app_y}	
	\end{align}
 	In challenging heterophilic graphs, this may produce better results in virtue of its better enhancement of locality information from input graph, which will be demonstrated in experimental sections.
	
	During the training stage, we feed the mixed features $ \mathbf{H}=\{\mathbf{h}_i \}_{i=1}^{|\mathcal{V}| } $ (here we drop the superscript $ K $ without causing confusion) and the adjacency matrix $ \mathbf{A} $ to GNN to predict labels. As is shown in Eq. \eqref{eq:mixup_loss}, the loss function consists of two parts. For labeled nodes, we use the ground truth labels as supervision signals. While for unlabeled nodes, we use the mixed labels for guidance. $ \lambda $ is a hyper-parameter used to balance the influence of mixed labels $ \bar{\mathbf{Y}} $. At the inference stage, we don't perform Mixup and the GNN accepts the original features and adjacency matrix as input.
	\begin{align}
		\mathcal{L}=\sum_{v\in \mathcal{V}_l }\ell(f(\mathbf{A}, \mathbf{H})_v , \mathbf{y}_v )+ \lambda \sum_{v\in \mathcal{V}_u }\ell(f(\mathbf{A}, \mathbf{H})_v , \bar{\mathbf{y}}_v ). \label{eq:mixup_loss}
	\end{align}
	
	\textbf{Relationship with Structure Learning.}
	The above Mixup is in some extent linked to graph structure learning, which helps GNN learning by optimizing the given graph structure to meet some desirable properties such as smooth node features and connectivity \citep{0022WZ20}. In Mixup, however, we modify the node features instead of the graph structure. In this sense, the training procedure after incorporating Mixup can be considered as a bilevel optimization problem. The upper-level optimization task treats the GNN $ f $ as the decision variable and aims to  
	minimize the label prediction loss in Eq. \eqref{eq:mixup_loss},
	while the lower-level optimization task is to minimize a regularization function  that regularizes the learned graph by modifying node features $ \mathbf{H} $ and labels $ \bar{\mathbf{Y}} $, which we will explain next.
	
	As a theoretical intuition and justification, Mixup operation \eqref{eq:geomix_gcn_x} and \eqref{eq:geomix_app_x} are gradient descent steps of two separate regularization functions which assess the quality of the mixed node features. Assume node features $ \mathbf{H}^{(t)} $ to be a continuous function over $ t \geq 0 $ with $ \mathbf{H}^{(0)}=\mathbf{X} $, we have the following theorem:
	\begin{theorem} \label{th:mixup_gcn}
		Mixup in Eq. \eqref{eq:geomix_gcn_x} and \eqref{eq:geomix_app_x} correspond to gradient descent steps of regularization functions $ F_1(\mathbf{H}; \mathbf{H}^{(t)}) $ and $ F_2(\mathbf{H}; \mathbf{H}^{(0)}) $.
		\begin{align}
			F_1(\mathbf{H}; \mathbf{H}^{(t)})= \sum_{u \in \mathcal{V}} \|\mathbf{h}_u - \mathbf{h}_u^{(t)}\|_2^2 +\beta \sum_{(u,v) \in \mathcal{E}} e_{uv} \|\mathbf{h}_u - \mathbf{h}_v\|_2^2 , \label{eq:gcn_mixup_cost} \\
			F_2(\mathbf{H}; \mathbf{H}^{(0)})= \sum_{u \in \mathcal{V}} \|\mathbf{h}_u - \mathbf{h}_u^{(0)}\|_2^2 +\beta \sum_{(u,v) \in \mathcal{E}} e_{uv} \|\mathbf{h}_u - \mathbf{h}_v\|_2^2 , \label{eq:app_mixup_cost} 
		\end{align}
		where $ \beta $ is related to $ \alpha $.
	\end{theorem} 
	The first term in Eq. \eqref{eq:gcn_mixup_cost} and \eqref{eq:app_mixup_cost} promote the proximity between the updated node feature and its current state or original state, while the second term encourages the similarity of features among neighboring nodes.
	We can obtain a similar cost function regarding mixed labels $ \bar{\mathbf{Y}} $, which operations in Eq. \eqref{eq:geomix_gcn_y} and \eqref{eq:geomix_app_y} serve to descend. 
	
	\textbf{Complexity analysis.}
	In both Mixup operations \eqref{eq:geomix_gcn_x} and \eqref{eq:geomix_app_x}, the computational complexity of calculating $ \mathbf{H}^{(t)} $ is $ O(|\mathcal{V}|F + |\mathcal{E}|F  ) $, where $ F $ is the number of input features. This is because the aggregating part involving $ e_{uv} $ can be implemented as a product of a sparse matrix with a dense matrix. Applying $ K $ consecutive Mixup operations multiplies the storage and time requirements by a factor of $ K $. In practice, $ K $ is usually 2. Similarly, the complexity of Mixup operation for labels is $ O(|\mathcal{V}|C + |\mathcal{E}|C )$, where $ C $ is the number of classes. As a result, the overall GNN training complexity after including Geometric Mixup remains consistent with conventional GNN training procedure.
	
	\subsection{Extending Geometric Mixup Beyond Vicinity: An Adaptive Approach}
	Geometric Mixup operations provided in previous sections have two limitations. Firstly, the aggregating weights $ e_{vu} $ are non-parametric, i.e., they are determined solely by adjacency matrix and need no training. Consequently, inappropriate weights may be assigned when the graph structure contains noise. Secondly, restriction imposed by graph structure greatly reduces Mixup choices, since a node can never have a chance to be mixed with a distant node.
	
	To address the aforementioned limitations, we allow a node to be mixed with any other node and adaptively learn the aggregating weights, as shown in Eq. \eqref{eq:geomix_diff_x}.
	\begin{align}
		\begin{aligned}
		 \hat{\mathbf{h}}^{(t+1)}_v&=\alpha \mathbf{h}^{(t)}_v +(1-\alpha) 	\sum_{u,v \in \mathcal{V} } a_{vu}^{(t)}\mathbf{h}^{(t)}_u \\
			\mathbf{h}^{(t+1)}_v&=(1-\eta)\hat{\mathbf{h}}^{(t+1)}_v+ \eta \sum_{u \in {\mathcal{N}}(v) } e_{vu} \mathbf{h}^{(t)}_u.
		\end{aligned}
 \label{eq:geomix_diff_x} 		
	\end{align}
	where $ a_{uv}^{(t)} $ is a time-frame-specific weight given by a weight prediction module which we will specify later.  $ e_{uv} $ is given by common normalized adjacency matrix. $ \eta $ is a hyper-parameter specifying the weight of adaptive all-pair aggregating results. 
	Symmetrically, the Mixup operation for labels is
	\begin{align}
	    \begin{aligned}
		\hat{\bar{\mathbf{y}}}^{(t+1)}_v&=\alpha \mathbf{y}^{(t)}_v+(1-\alpha)\sum_{u,v \in \mathcal{V} } a_{vu}^{(t)} \bar{\mathbf{y}}^{(t)}_u  \\
		\bar{\mathbf{y}}^{(t+1)}_v&=(1-\eta)\hat{\bar{\mathbf{y}}}^{(t+1)}_v+ \eta \sum_{u \in {\mathcal{N}}(v) } e_{vu} \bar{\mathbf{y}}^{(t)}_u.  
		\end{aligned}
		\label{eq:geomix_diff_y} 
	\end{align}
	
	To adaptively and efficiently predict the aggregating weight $ a_{vu}^{(t)} $, we adopt a simple project-then-dot-product method, as displayed in Eq. \eqref{eq:agg_weight_predict}.
	\begin{align}
		a_{vu}^{(t)} = \frac{ ( \mathbf q_v^{(t)} )^\top   \mathbf k_u^{(t)}  }{ \sum_{w\in \mathcal{V}} ( \mathbf q_v^{(t)} )^\top   \mathbf k_w^{(t)}   } ,  \label{eq:agg_weight_predict}
	\end{align}
	with
	\begin{align}
		\mathbf{q}_v^{(t)} = \frac{\mathbf{h}_v^{(t)} \mathbf{W}_q^{(t)} }{\| \mathbf{h}_v^{(t)} \mathbf{W}_q^{(t)} \|_2}, 
		\quad \mathbf{k}_u^{(t)} = \frac{\mathbf{h}_u^{(t)} \mathbf{W}_k^{(t)} }{\| \mathbf{h}_u^{(t)} \mathbf{W}_k^{(t)} \|_2}, 
	\end{align}
	where $ \mathbf{W}_q^{(t)} \in \mathbb{R}^{F\times F'} $ and $ \mathbf{W}_k^{(t)}  \in \mathbb{R}^{F\times F'} $ ($ F $ and $ F' $ is the dimension of input features and hidden features) are two learnable projection matrices. Eq. \eqref{eq:agg_weight_predict} aligns with the self-attention mechanism of the Transformer \citep{transformer}. In this paradigm, each node can potentially mix its features/labels with those of any other node whose projected feature is similar. This addresses the limitation of having only a few choices of nodes for Mixup in previous Geometric Mixup. Moreover, the trainable parameters $ \mathbf{W}_q^{(t)} $ and $ \mathbf{W}_k^{(t)} $ can remedy the problem of inappropriate aggregating weights assigned by input graph.
	
	\textbf{Complexity analysis.}
	The all-pair aggregating operation guided by weight $ a_{vu}^{(t)} $ in Eq. \eqref{eq:geomix_diff_x} can be written in the following matrix form
	\begin{align}
		\mathbf{M}^{(t)}=\left( \operatorname{diag} \left(\mathbf{Q}^{(t)}(\mathbf{K}^{(t)} )^{\top} \mathbf{1} \right)	\right)^{-1} \left( \mathbf{Q}^{(t)}(\mathbf{K}^{(t)} )^{\top} \right) \mathbf{H}^{(t)}, 
	\end{align}
	where $ \mathbf{Q}^{(t)} $ and $ \mathbf{K}^{(t)} $ are constructed by concatenating $ \mathbf{q}^{(t)}_u, u\in \mathcal{V} $ and $ \mathbf{k}^{(t)}_u, u\in \mathcal{V} $ vertically respectively. 
	By using the associative law of matrix multiplication, the above equation is equivalent to
	\begin{align}
		\mathbf{M}^{(t)}=\left( \operatorname{diag}\left(\mathbf{Q}^{(t)} \left( (\mathbf{K}^{(t)} )^{\top} \mathbf{1} \right) \right)	\right)^{-1}  \mathbf{Q}^{(t)} \left( (\mathbf{K}^{(t)} )^{\top}  \mathbf{H}^{(t)} \right). \label{eq:geomix_diff_x_rearange}
	\end{align}
	By first calculating $ (\mathbf{K}^{(t)})^{\top} \mathbf{1} $ and 
	$ (\mathbf{K}^{(t)} )^{\top}  \mathbf{H}^{(t)} $ rather than 
	$ \mathbf{Q}^{(t)}(\mathbf{K}^{(t)} )^{\top} $, we reduce the quadratic complexity to linear w.r.t the number of nodes. 
	The time complexity of Eq. \eqref{eq:geomix_diff_x_rearange} is $ O(|\mathcal{V}| FF' ) $ where $ F $ and $ F' $ are dimensions of input features and hidden features. Combining this with the complexity analysis in Sec. \ref{sec:geomix}, the overall complexity of Mixup in Eq. \eqref{eq:geomix_diff_x} is $ O(|\mathcal{V}| FF'+ (|\mathcal{V}|+|\mathcal{E}| )F) $.
	Through a similar analysis, the time complexity of label Mixup in Eq. \eqref{eq:geomix_diff_y} is $ O(|\mathcal{V}| CF'+ (|\mathcal{V}|+|\mathcal{E}| )C ) $, where $ C $ is the number of classes. Therefore, this variant of Geometric Mixup still preserves the complexity order of conventional GNN training.

	\section{Experiments}
	In this section, we conduct comprehensive experiments to evaluate Geometric Mixup on an extensive set of node classification datasets. Specifically, we focus on the following research questions:
	
	\noindent\bb\;\textbf{1)} Can Geometric Mixup consistently and significantly improve the performance of GNNs on common benchmarks with limited labeled data? Additionally, can it cope with both homophily (where adjacent nodes tend to share similar labels) \cite{lim2021new} and heterophily (which means adjacent nodes tend to have different labels)? 
	
	\noindent\bb\;\textbf{2)} Can Geometric Mixup consistently and significantly enhance the ability of GNNs in out-of-distribution (OOD) generalization tasks? That is, one has access to limited distributions in training set and needs to generalize to datasets from distributions different from those of the training data.
	
	\noindent\bb\;\textbf{3)} Are the proposed components in Geometric Mixup effective and necessary for the achieved performance?
	
	\noindent\bb\;\textbf{4)} Can Geometric Mixup help GNNs learn more discriminative representations for improved class differentiation?
	
	\paragraph{Implementation details}
	We implement the three proposed Geometric Mixup methods defined in Eq. \eqref{eq:geomix_gcn_x}, \eqref{eq:geomix_app_x}, and \eqref{eq:geomix_diff_x}, naming them \ours-I, \ours-II, and \ours-III, respectively.
	Unless otherwise specified, we employ GCN \cite{gcn} as the foundational GNN for both Geometric Mixup and other competing methods that utilize a GNN backbone. Following an optimization of architecture-related hyperparameters for the standard GCN, which includes the number of layers and hidden size, we adopt the same architecture for Geometric Mixup to ensure a fair comparison.
	To reduce the number of hyper-parameters, we set $ \lambda $ in Eq. \eqref{eq:mixup_loss} to a default value of 1 except in a few cases, since this consistently produces exemplary results across a wide range of test cases. 
	For additional implementation and hyper-parameter details, please refer to the appendix.
	
		\begin{table*}[tb!]
		\caption{Mean and standard deviation (with five runs using random initializations) of testing accuracy on node classification benchmarks.}
		\label{tab:common_datasets}
		\begin{tabular}{lccccccc}
			\toprule
			\textbf{Methods} & \textbf{Cora} & \textbf{CiteSeer} & \textbf{PubMed} & \textbf{CS} & \textbf{Physics} & \textbf{Squirrel} & \textbf{Chameleon} \\
			\midrule
			GCN & 81.63 ± 0.45 & 71.64 ± 0.33 & 78.88 ± 0.65 & 91.16 ± 0.52 & 92.85 ± 1.03 & 39.47 ± 1.47 & 41.32 ± 3.22 \\
			GAT & 82.98 ± 0.88 & 72.20 ± 0.99 & 78.58 ± 0.52 & 90.57 ± 0.37 & 92.70 ± 0.58 & 35.96 ± 1.73 & 39.29 ± 2.84 \\
			SGC & 80.35 ± 0.24 & 71.87 ± 0.14 & 78.75 ± 0.17 & 90.37 ± 1.01 & 92.80 ± 0.15 & 39.04 ± 1.92 & 39.35 ± 2.82 \\
			APPNP & 83.33 ± 0.52 & 71.83 ± 0.52 & 79.78 ± 0.66 & 91.97 ± 0.33 & 93.86 ± 0.33 & 37.64 ± 1.63 & 38.25 ± 2.83 \\
			GloGNN & 82.31 ± 0.42 & 72.16 ± 0.64 & 78.95 ± 0.42 & 90.82 ± 0.45 & 92.79 ± 0.67 & 35.77 ± 1.32 & 40.13 ± 3.91 \\
			\midrule
			Mixup & 81.84 ± 0.94 & 72.20 ± 0.95 & 79.16 ± 0.49 & 91.36 ± 0.37 & 93.89 ± 0.49 & 37.95 ± 1.52 & 39.56 ± 3.13 \\
			GraphMixup & 82.16 ± 0.74 & 72.13 ± 0.86 & 78.82 ± 0.52 & 91.27 ± 0.55 & 93.62 ± 0.41 & 37.84 ± 1.46 & 39.82 ± 2.35 \\
			GraphMix & 83.80 ± 0.62 & 74.28 ± 0.45 & 79.38 ± 0.39 & 91.89 ± 0.36 & 94.32 ± 0.28 & 38.41 ± 1.36 & 41.75 ± 3.51 \\
			\midrule
			\ours-I & \textbf{84.08 ± 0.74} & \textbf{75.06 ± 0.36} & \textbf{80.06 ± 0.93} & 92.13 ± 0.06 & \textbf{94.51 ± 0.07} & \cellcolor{lightgrey}\textbf{40.95 ± 1.12} & 41.94 ± 3.41 \\
			\ours-II & 83.94 ± 0.50 & \cellcolor{lightgrey}\textbf{75.12 ± 0.26} & 79.98 ± 0.35 & \textbf{92.14 ± 0.11} & \cellcolor{lightgrey}\textbf{94.56 ± 0.06} & 40.75 ± 1.30 & \cellcolor{lightgrey}\textbf{42.67 ± 2.44} \\
			\ours-III & \cellcolor{lightgrey}\textbf{84.22 ± 0.85} & 73.60 ± 0.83 & \cellcolor{lightgrey}\textbf{80.18 ± 0.99} & \cellcolor{lightgrey}\textbf{92.23 ± 0.14} & 94.34 ± 0.04 & \textbf{40.78 ± 1.75} & \textbf{42.58 ± 3.38} \\
			\bottomrule
		\end{tabular}
	\end{table*}

	\paragraph{Competitors}
	We mainly compare with GCN \cite{gcn}, the GNN backbone of Geometric Mixup, for testing the efficacy of Geometric Mixup.  We also compare with several state-of-the-art Mixup methods for node classification: Mixup \cite{wang2021mixup}, GraphMix \cite{verma2021graphmix} and GraphMixup \cite{wu2022graphmixup}. Furthermore, we compare with more advanced GNNs: GAT \cite{gat}, SGC \cite{sgc}, APPNP \cite{appnp} and GloGNN \cite{glognn}. In OOD generalization tasks, we add standard empirical risk minimization (ERM), DANN \cite{ganin2016domain}, EERM \cite{wu2022handling} as competitive methods.

	\subsection{Common Node Classification Datasets}
	We first conduct experiments on several commonly used graph datasets, including three citation networks \texttt{Cora}, \texttt{CiteSeer} and \texttt{PubMed} \cite{planetoid}; two co-authorship networks: \texttt{CS} and \texttt{Physics} \cite{shchur2018pitfalls}; and two heterophilic graphs: \texttt{Squirrel} and \texttt{Chameleon} \cite{platonov2023a}, where neighboring nodes tend to have distinct labels. For citation networks, we use the same data splits as in \cite{planetoid}, which selects 20 nodes from each class as training set, 1,000 nodes in total as validation set and 500 nodes as test set. For two co-authorship networks,  we follow the splits in \cite{shchur2018pitfalls}, i.e.,  20 labeled nodes per class as the training set, 30 nodes per class as the validation set, and the rest as the test set. For the two heterophilic datasets, we follow the recent paper \cite{platonov2023a} that filters out the overlapped nodes in the original datasets and use its provided data splits. 
	
    As displayed in Table \ref{tab:common_datasets}, all three variants—\ours-I, \ours-II, and \ours-III—significantly enhance the performance of GCN, their foundational GNN architecture, across all datasets.  
	In Comparison to other advanced GNNs, they consistently achieve superior accuracy even using simple GCN as the GNN backbone.
	Furthermore, each of the three proposed Geometric Mixup variants consistently outperforms state-of-the-art Mixup competitors. 
	These results suggest that leveraging geometry information for Mixup is highly effective in improving the performance of GNN and addressing challenges caused by limited labeled data. It is likely to yield superior results compared to Mixup that randomly pairs nodes.
	
	\begin{table}[tb!]
		\caption{Mean and standard deviation of testing accuracy for OOD generalization on Twitch-FR, Twitch-PT and Twitch-RU. All methods use GCN as foundational GNN architecture.}
		\label{tab:twitch_ood}
		\begin{tabular}{lccc}
			\toprule
			\textbf{Mothods} & \textbf{Twitch-FR} & \textbf{Twtich-PT} & \textbf{Twitch-RU} \\
			\midrule
			ERM & 57.98 ± 2.41 & 64.58 ± 0.63 & 59.74 ± 3.89 \\
			EERM & 58.03 ± 0.53 & 65.90 ± 0.47 & 59.71 ± 1.87 \\
			DANN & 52.16 ± 5.29 & 64.92 ± 1.17 & 61.36 ± 4.09 \\
			\midrule
			Mixup & 54.01 ± 2.23 & \textbf{66.31 ± 0.65} & 57.32 ± 2.29 \\
			GraphMix & 56.62 ± 2.36 & 65.22 ± 0.63 & 65.73 ± 1.29 \\
			\midrule
			\ours-I & 57.82 ± 2.61 & \cellcolor{lightgrey}\textbf{66.96 ± 0.63} & 64.00 ± 1.62 \\
			\ours-II & \cellcolor{lightgrey}\textbf{61.67 ± 1.38} & 65.58 ± 1.46 & \cellcolor{lightgrey}\textbf{69.95 ± 2.67} \\
			\ours-III & \textbf{60.97 ± 2.14} & 65.38 ± 1.89 & \textbf{69.55 ± 2.46} \\
			\bottomrule
		\end{tabular}
	\end{table}

	\subsection{Handling Distribution Shifts in Unseen Domains}
	We proceed to test Geometric Mixup's capability of handling distribution shifts in OOD generalization tasks. We conduct experiments on \texttt{Twitch-explicit} dataset, which contains multiple networks where Twitch users are nodes, and mutual friendships between them are edges \cite{rozemberczki2021multi}. Since each graph is associated to users of a particular region, distribution shifts occur between different graphs. We train and validate our model on three graphs: \texttt{DE}, \texttt{EN}, \texttt{ES}, and perform a random split into 50\% training, 25\% validation, and 25\% in-distribution-test sets. After training, we directly evaluate the model on \texttt{FR}, \texttt{PT} and \texttt{RU} datasets. 
	
	To make our experiments solid, we not only compare our methods with state-of-the-art Mixup methods for node classification, but also include comparisons with EERM and DANN, two of the most advanced methods designed to address distribution shifts. As GraphMixup contains an edge prediction module, which relies on domain knowledge and cannot handle distribution shifts effectively, we do not include it in this section.
	We report the results in Table \ref{tab:twitch_ood}. The three Geometric Mixup variants substantially enhance the performance over ERM, the most basic OOD training approach. Notably, on \texttt{Twitch-RU}, the relative improvement reaches a remarkable 17.09\%. Furthermore, Geometric Mixup shows superior performance over other advanced methods across all three datasets, thus validating the efficacy of our design in improving the generalization capabilities of the underlying GNN. As indicated in \cite{lim2021new}, the graphs within \texttt{Twitch-explicit} exhibit heterophilic characteristics. Consequently, experiments in this section also underscore Geometric Mixup's effectiveness in handling heterophilic graphs. Moreover, the results presented in Table \ref{tab:twitch_ood} clearly indicate that \ours-II outperforms \ours-I on such graphs. This observation substantiates the assertions made in Section \ref{sec:geomix}.
	
	\begin{figure*}[tb!]
		\centering
		\includegraphics[width=0.9\linewidth]{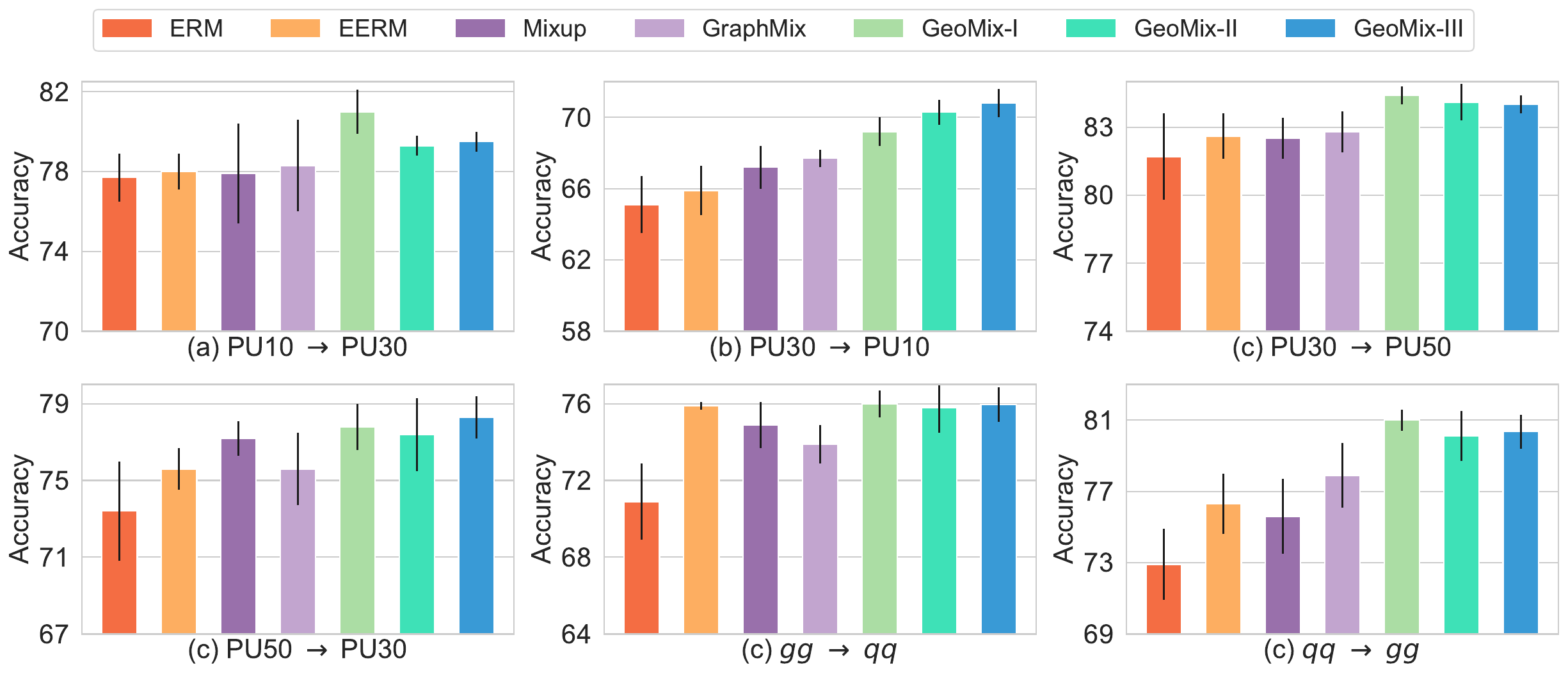}
		\caption{Mean testing accuracy and standard deviation of generalization task in \texttt{Pileup Mitigation} dataset with different PU conditions and physical processes. Expressions like PU10 $ \rightarrow $ PU30 represent PU condition shifts. $ gg\ \rightarrow \ qq $ and $ qq\ \rightarrow \ gg $ indicate physical processes shifts. }
		\label{fig:pileup_res}
	\end{figure*}
	
	\subsection{OOD Generalization in High Energy Physics}
	Next, we test Geometric Mixup in OOD generalization using \texttt{Pileup Mitigation} dataset from the realm of High Energy Physics (HEP) \cite{liu2023structural}. It comprises multiple graphs, with each corresponding to a beam of proton-proton collisions. The nodes in each graphs represent particles generated by these collisions in the Large Hadron Collider, categorized into primary collisions (LC) and nearby bunch crossings (OC). Node features encode various physics characteristics of these particles.
	Graphs are constructed from input features using KNN method \cite{liu2023structural}.
	The task is to identify whether a neutral particle is from LC or OC. 
	The distribution shifts can be attributed to two sources: first, variations in pile-up (PU) conditions, such as generalization from PU10 to PU30; second, changes in the types of the particle decay, for example, generalization from $ pp $ to $ qq $.
	To establish a semi-supervised learning setting, for each generalization task, we choose 10 graphs from the source domain and randomly allocate 20\% of neutral nodes (particles) as the training set, 80\% forming the validation set. For the target domain, we use 20 graphs and test the model on all the neutral nodes. 
	
	This task presents a twofold challenge. Firstly, it necessitates an in-depth understanding of complex domain knowledge within the HEP field. Secondly, it involves conditional structure shifts, a new type of challenging distribution shift identified by \cite{liu2023structural}. The results are presented in Fig. \ref{fig:pileup_res}. 
	Despite the substantial challenges, \ours-I, \ours-II and \ours-III all significantly enhance the testing accuracy of the underlying GCN across all tasks. Notably, the most substantial improvements are observed in distribution shifts caused by different physical processes, which are more demanding than shifts arising from variations in PU conditions \cite{liu2023structural}. In $ gg\ \rightarrow \ qq $  and $ qq\ \rightarrow \ gg $, Geometric Mixup results in relative improvements as high as 7.22\% and 11.14\% over ERM. Additionally, Geometric Mixup consistently outperforms other advanced competitors throughout all scenarios.
	These findings demonstrate that with the aid of Geometric Mixup, GNNs can effectively acquire complex scientific knowledge from limited training data to address real-world challenges. They also highlight Geometric Mixup's capability of addressing distribution shifts between source and target graphs.
	
	\subsection{Image and Text Classification with Low Label Rates}
	
	\begin{table*}[tb!]
		\caption{Testing accuracy on image (\texttt{STL10} and \texttt{CIFAR10}) and text (\texttt{20News}) classification. The second column displays the number of samples per class in the training set.}
		\label{tab:image_text}
		\resizebox{\linewidth}{!}{
		\begin{tabular}{| l | c | cccc | ccc | ccc|}
			\hline
			\textbf{Dataset} & \specialcell{\textbf{\#samples}\\\textbf{per class}}  & MLP & GCN & GAT & SGC & Mixup & GraphMixup & GraphMix & \ours-I & \ours-II & \ours-III \\
			\hline
			\multirow{2}{*}{\textbf{STL10}} & 10 & 66.6 ± 0.8 & 67.3 ± 0.4 & 67.1 ± 0.6 & 66.6 ± 0.2 & 67.7 ± 0.8 & 67.0 ± 0.7 & 67.7 ± 0.7 & \textbf{68.3 ± 0.6} & 68.2 ± 0.6 & \cellcolor{lightgrey}\textbf{68.4 ± 0.7} \\
			& 20 & 70.0 ± 0.6 & 69.7 ± 0.4 & 69.5 ± 0.4 & 68.9 ± 0.2 & 69.9 ± 0.6 & 69.7 ± 0.4 & 70.5 ± 0.5 & 70.5 ± 0.2 & \textbf{70.8 ± 0.2} & \cellcolor{lightgrey}\textbf{70.9 ± 0.4} \\
			\hline
			\multirow{2}{*}{\textbf{CIFAR10}} & 10 & 68.8 ± 0.7 & 70.0 ± 0.8 & 70.1 ± 0.8 & 69.6 ± 0.7 & 69.3 ± 1.1 & 69.7 ± 1.4 & 69.8 ± 1.3 & \textbf{71.1 ± 1.6} & 71.1 ± 1.1 & \cellcolor{lightgrey}\textbf{71.3 ± 0.6} \\
			& 20 & 72.0 ± 0.6 & 71.8 ± 0.5 & 71.7 ± 0.6 & 71.9 ± 0.5 & 72.0 ± 0.5 & 71.6 ± 0.6 & 72.3 ± 0.2 & \textbf{72.9 ± 0.4} & 72.9 ± 0.4 & \cellcolor{lightgrey}\textbf{73.2 ± 0.2} \\
			\hline
			\multirow{2}{*}{\textbf{20News}} & 100 & 55.9 ± 0.3 & 56.6 ± 0.3 & 56.9 ± 0.5 & 55.6 ± 0.8 & 57.5 ± 0.4 & 57.8 ± 0.6 & 57.5 ± 0.2 & \cellcolor{lightgrey}\textbf{58.6 ± 0.2} & 58.2 ± 0.1 & \textbf{58.5 ± 0.1} \\
			& 200 & 60.0 ± 0.3 & 60.4 ± 0.6 & 60.8 ± 0.4 & 59.2 ± 0.3 & 60.9 ± 0.6 & 60.9 ± 0.5 & 61.0 ± 0.7 & \cellcolor{lightgrey}\textbf{61.5 ± 0.4} & \textbf{61.4 ± 0.3} & 61.3 ± 0.4 \\
			\hline
		\end{tabular}
	}
	\end{table*}
	
	We extend our experiments to include the \texttt{STL10}, \texttt{CIFAR10}, and \texttt{20News} datasets to evaluate Geometric Mixup's performance in standard classification tasks with limited labeled data. In \texttt{20News} provided in \cite{pedregosa2011scikit}, we select 10 topics and use words with a TF-IDF score exceeding 5 as features. For \texttt{STL10} and \texttt{CIFAR10}, both image datasets, we initially employ the self-supervised  approach SimCLR \cite{chen2020simple}, which does not use any labels for training, to train a ResNet-18 model for extracting feature maps used as input features. Since these datasets lack inherent graphs, we utilize the KNN method to construct input graphs. We leave more details in the appendix.
	
	The results are presented in Table \ref{tab:image_text}. Notably, all three Geometric Mixup methods consistently outperform GCN, their underlying GNN, as well as other GNNs across all cases. Furthermore, they achieve superior results compared to three state-of-the-art Mixup competitors. These findings underscore the broad applicability of Geometric Mixup, spanning not only graph-structured datasets but also image and text classifications where explicit graphs are absent.
	
	\subsection{Ablation Study}
	\begin{table}[tb!]
		\caption{Results of the ablation studies on \texttt{Cora}, \texttt{CiteSeer} and \texttt{Squirrel}. In ``Random Mix", we randomly pair nodes to perform Mixup. In ``w/o Locality", we remove the locality enhancement part in \ours-I. $ \Delta_{\ours-I} $ represents the relative performance degradation compared to \ours-I. }
		\label{tab:ablatioin}
            \begin{tabular}{lccc}
            \toprule
            \textbf{Method}            & \textbf{Cora} & \textbf{CiteSeer} & \textbf{Squirrel} \\
            \midrule
            GCN & 81.6 ± 0.5    & 71.6 ± 0.3        & 39.5 ± 1.5        \\
            \ours-I  & 84.1 ± 0.7    & 75.1 ± 0.4        & 41.0 ± 1.1       \\
            \midrule
            Random Mix & 82.1 ± 0.5    & 69.8 ± 1.3        & 37.8 ± 1.3        \\
            $ \Delta_{\ours-I} $ & (-2.38\%)     & (-7.06\%)         & (-7.80\%)         \\
            \midrule
            w/o Locality & 84.0 ± 0.6    & 73.7 ± 0.8        & 38.3 ± 1.6        \\
            $ \Delta_{\ours-I} $  & (-0.12\%)     & (-1.86\%)         & (-6.59\%) \\
            \bottomrule
            \end{tabular}
	\end{table}
	
	In this section, we conduct ablation studies to demonstrate the efficacy and necessity of certain design choices in Geometric Mixup. Firstly, we aim to assess the improvements brought about by the utilization of geometry information in Mixup.  To achieve this, we randomly pair nodes for Mixup while keeping all other aspects of the training pipeline consistent with Geometric Mixup. Secondly, we seek to understand the effects of locality enhancement, so we remove the locality enhancement part in \ours-I and keep the other design elements unchanged. 
	
	We conduct these experiments on \texttt{Cora}, \texttt{CiteSeer} and \texttt{Squirrel}, with the latter being a heterophilic graph. The results are displayed in Table \ref{tab:ablatioin}. There is a substantial drop in performance when we do not incorporate geometry information. One possible explanation is that randomly mixing nodes can introduce unwanted external noise into each node's receptive field, thus negatively affecting the accuracy of information exchange during message passing.
	
	After disabling the locality enhancement, noticeable performance declines are observed in \texttt{CiteSeer} and \texttt{Squirrel}, while no significant difference is observed in \texttt{Cora}. These outcomes can be attributed to homophily. As analyzed in Sec. \ref{sec:gcn_mixup}, a reduction in homophily can cause the basic geometry-aware Mixup without locality enhancement to inadequately preserve locality information and ensure the diversity of synthetic data, thereby diminishing the efficacy of Mixup. According to \cite{lim2021new}, even though \texttt{Cora} and \texttt{CiteSeer} are homophilic graphs, \texttt{CiteSeer} exhibits a lower homophily ratio. Consequently, there is a more pronounced performance drop in \texttt{CiteSeer} compared to \texttt{Cora}. In the case of the heterophilic graph \texttt{Squirrel}, the performance drop is even more substantial, reaching 6.59\%. These results substantiate the necessity of locality enhancement for Geometric Mixup.

	\subsection{Visualization}
	
 \begin{figure}[tb!]
		\centering
		\begin{subfigure}[b]{0.4\linewidth}
			\includegraphics[width=\linewidth]{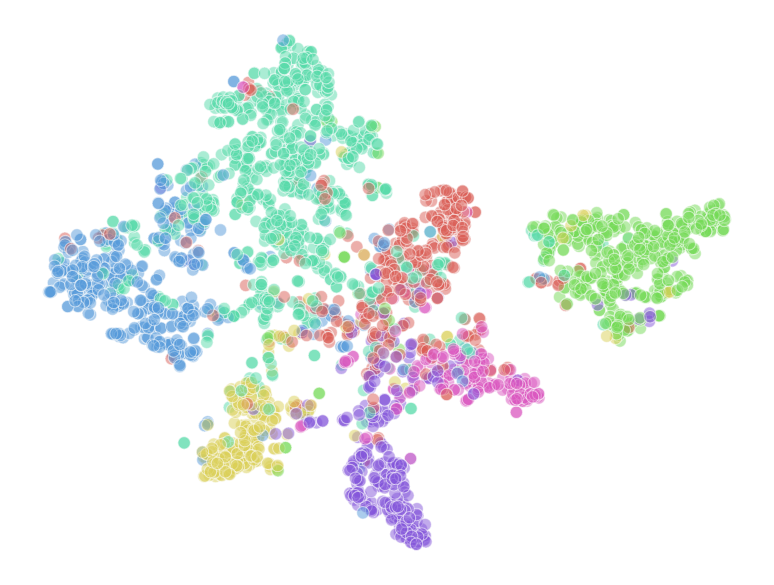}
			\caption{GCN on Cora}
		\end{subfigure}
		\begin{subfigure}[b]{0.4\linewidth}
			\includegraphics[width=\linewidth]{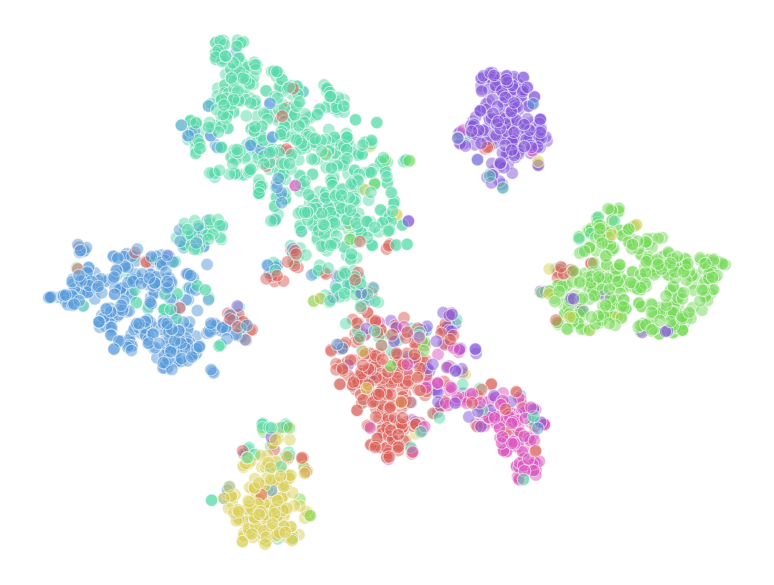}
			\caption{\ours-I on Cora}
		\end{subfigure}
		\begin{subfigure}[b]{0.4\linewidth}
			\includegraphics[width=\linewidth]{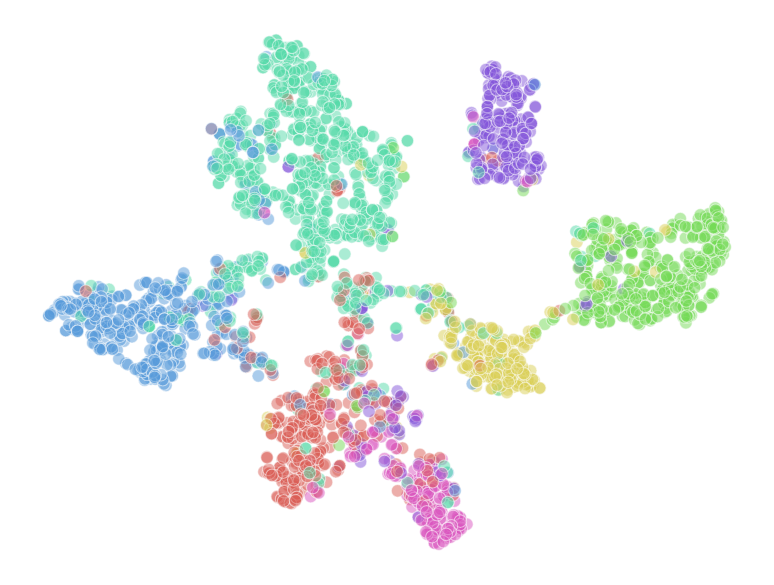}
			\caption{\ours-II on Cora}
		\end{subfigure}
		\begin{subfigure}[b]{0.4\linewidth}
			\includegraphics[width=\linewidth]{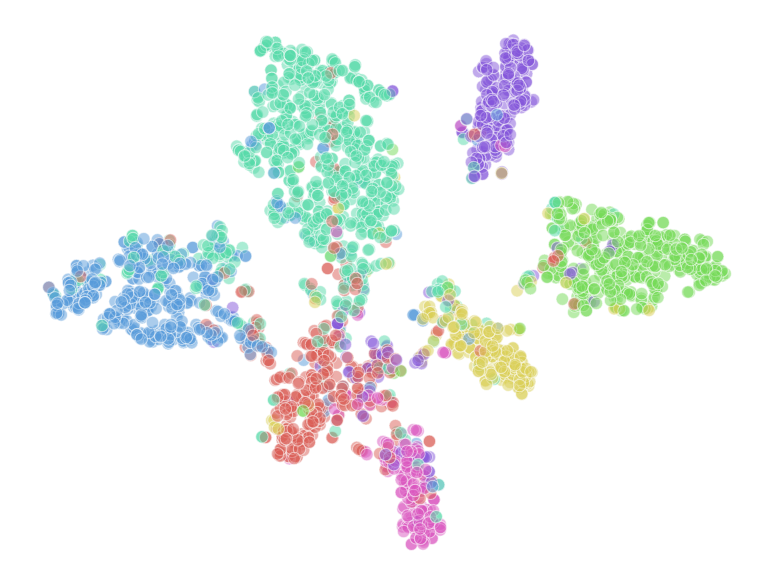}
			\caption{\ours-III on Cora}
		\end{subfigure}
		\caption{The learned representations of the nodes in the Cora datasets by GCN and Geometric Mixup. Colors denote the ground-truth class labels.}
		\label{fig:tsne_cora}
	\end{figure}
	
	\begin{figure}[tb!]
		\centering
		\begin{subfigure}[b]{0.4\linewidth}
			\includegraphics[width=\linewidth]{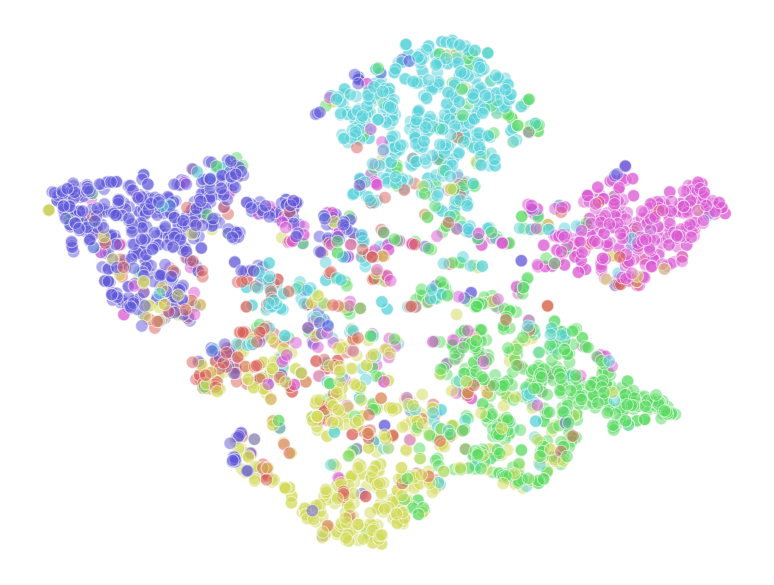}
			\caption{GCN on CiteSeer}
		\end{subfigure}
		\begin{subfigure}[b]{0.4\linewidth}
			\includegraphics[width=\linewidth]{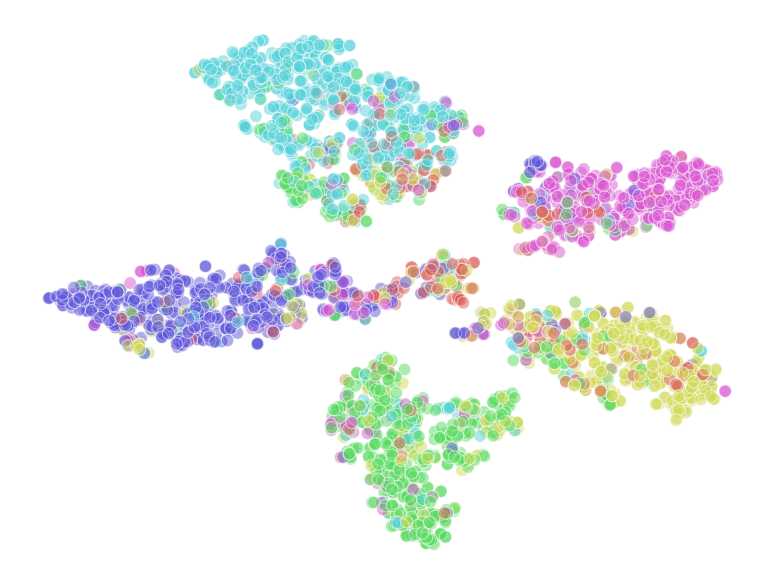}
			\caption{\ours-I on CiteSeer}
		\end{subfigure}
		\begin{subfigure}[b]{0.4\linewidth}
			\includegraphics[width=\linewidth]{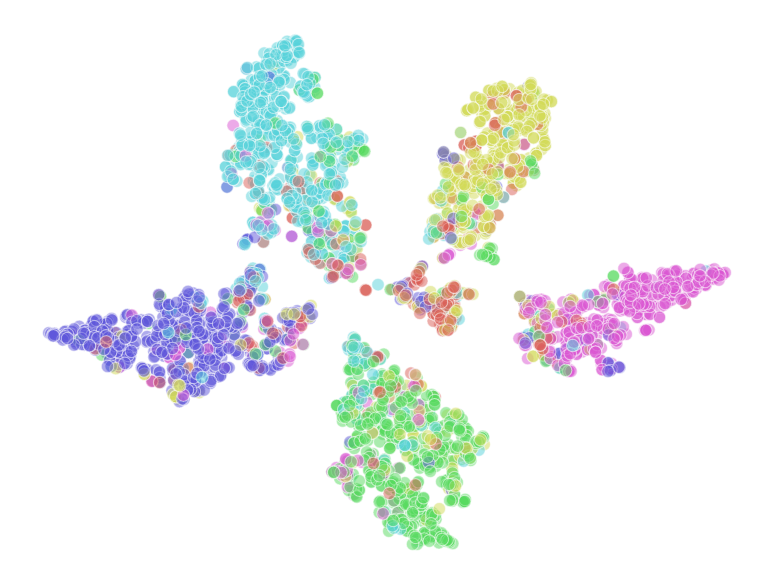}
			\caption{\ours-II on CiteSeer}
		\end{subfigure}
		\begin{subfigure}[b]{0.4\linewidth}
			\includegraphics[width=\linewidth]{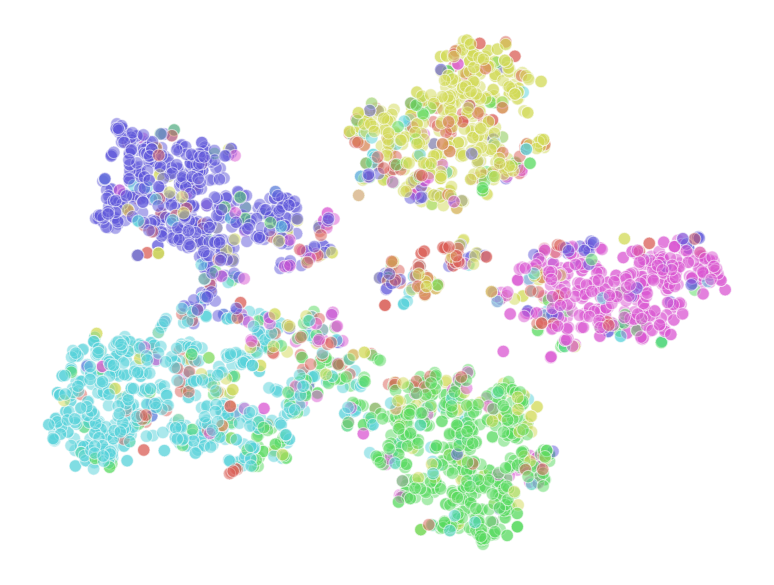}
			\caption{\ours-III on CiteSeer}
		\end{subfigure}		
	\caption{The learned representations of the nodes in the CiteSeer datasets by GCN and Geometric Mixup.}
	\label{fig:tsne_citeseer}
	\end{figure}

	Fig. \ref{fig:tsne_cora} and \ref{fig:tsne_citeseer} displays the final-layer node representations learned by GCN and \ours on the Cora and CiteSeer datasets, using T-SNE \cite{van2008visualizing}. The figures reveal that the hidden representations learned with Geometric Mixup are more discriminative and conducive to clustering, as nodes from the same class are more tightly clustered, while nodes from different classes are more distant from each other. These highly discriminative representations contribute to improved class predictions.
	
	\subsection{Results of Using Other GNN Backbones}
	
	\begin{figure}[tb!]
		\centering
		\includegraphics[width=0.7\linewidth]{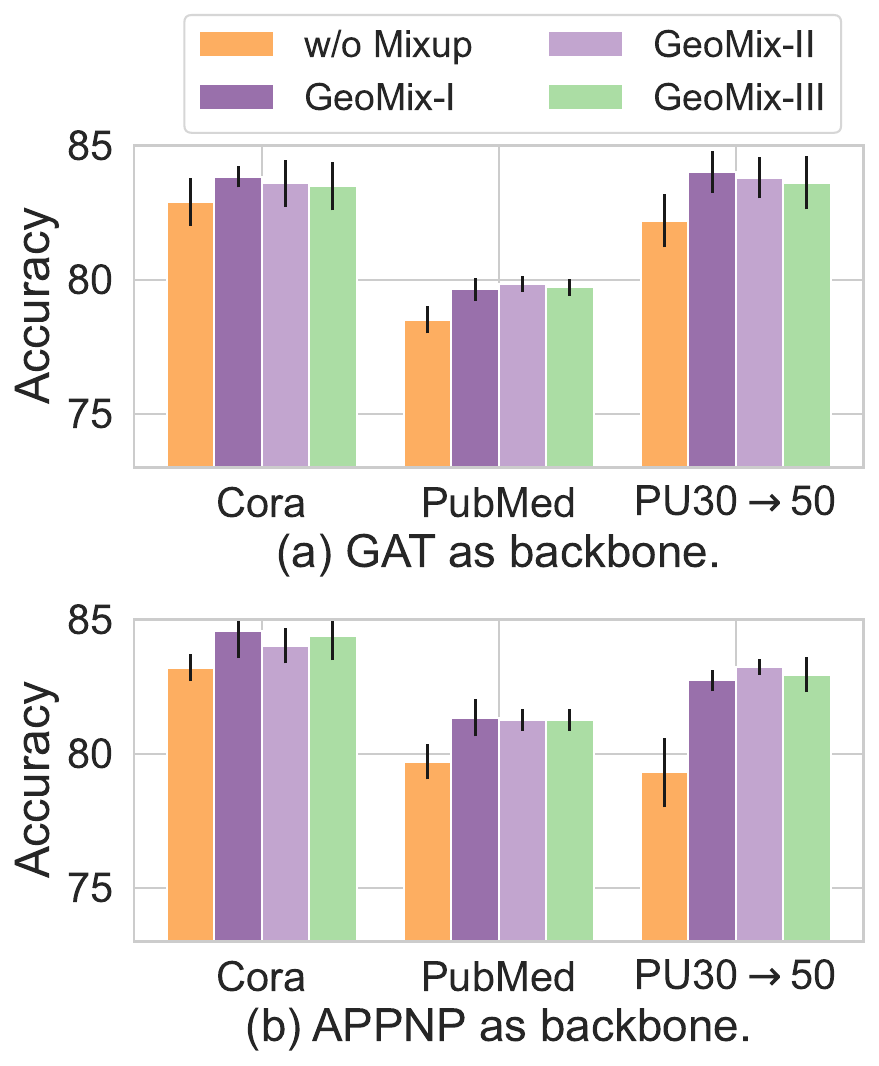}
		\caption {Results with other underlying GNN architectures.}
		\label{fig:backbone}
	\end{figure}

	In this section, we investigate the versatility of Geometric Mixup by altering the underlying GNN architectures. Specifically, we employ GAT \cite{gat} and APPNP \cite{appnp} as backbone GNNs and evaluate their performance with all three variants of Geometric Mixup. The results are presented in Fig. \ref{fig:backbone}. As shown, Geometric Mixup consistently enhances the performance of GAT and APPNP across both standard datasets and out-of-distribution (OOD) generalization tasks.

	\section{Related Works}
	
	\paragraph{Graph Neural Networks}
	Graph Neural Networks (GNNs) have become the de facto method for modeling graph-structured data. Among the various types of GNNs, message-passing-based approaches \cite{gcn, gat, appnp, sgc, xu2018representation} have gained prominence by defining graph convolutions through information propagation. These approaches generate the representation of a node by aggregating its own features along with those of its neighbors. Our work is orthogonal to them in that our model-agnostic Mixup operation serves as a data preprocessing step to enlarge the training set and broaden the data distribution, ultimately enhancing performance and generalization. 
	
	\paragraph{Mixup}
	As discussed in previous studies \cite{ZhangCDL18, tokozume2018between, verma2019manifold}, Mixup is a highly effective data augmentation technique that generates training samples through the interpolation of existing samples. However, Mixup is mostly used in image classification and has rarely been explored in graph learning tasks, particularly the node classification task.
	When considering node classification, while interpolating node features and labels to generate synthetic nodes seems intuitive,  the challenge lies in effectively establishing connections for these synthetic nodes. Care must be exercised during this process to avoid introducing excessive external noise into the information propagation mechanism, as it can detrimentally impact the performance of GNNs.
	In this domain, a few existing works either avoid explicitly connecting synthetic nodes \cite{wang2021mixup, verma2021graphmix} or introduce complex edge prediction modules \cite{wang2021mixup}. The former performs Mixup between layers of neural networks and tightly couples with the training model, potentially limiting its versatility. Conversely, the latter compromises on efficiency and generalization capability. To the best of our knowledge, our research marks the pioneering effort in integrating the graph geometry into Mixup operation. This integration allows for the construction of an explicit augmented graph, wherein synthetic nodes are systematically connected to relevant nodes. This approach enhances interpretability while maintaining the efficiency of the Mixup technique.
	
	\paragraph{Generalization on Graph Learning}
	Owing to the distribution shifts encountered between real-world testing and training data, there has been a growing emphasis on enhancing the capacity of GNNs to perform effectively on out-of-distribution (OOD) data. One line of work involves the application of adversarial training to promote the smoothness of the output distribution, such as BVAT \cite{deng2023batch} and GraphAT \cite{feng2019graph}. A more recent invariance learning approach, EERM \cite{wu2022handling}, introduces multiple context explorers, which are implemented as graph editors and are adversarially trained. 
    Another recent work \cite{graphglow} proposes learning a generalizable graph structure learner that can enhance the quality of the input graph structure when generalizing to unseen graphs, thereby improving the performance of the downstream GNN.
    However, it is worth noting that these methods introduce significant extra computational costs. In contrast, Geometric Mixup is more lightweight. It introduces only a few message-passing-based Mixup operations and operates in linear time with respect to the number of nodes and edges.
	
	\section{Conclusion}
	This paper proposes Geometric Mixup, a method leveraging geometry information for Mixup by interpolating features and labels with those from nearby neighborhood. We provide theoretic insights into our approach for utilizing graph structure and emphasizing the importance of enhancing locality information, a critical design aspect enabling our method to accommodate to both homophilic and heterophilic graphs. Additionally, we extend our strategy to facilitate all-pair Mixup and dynamically learn the mixing weights, overcoming the challenges posed by noise in the given graph structure. Our extensive experiments demonstrate that Geometric Mixup substantially improves the performance of underlying GNNs on both standard datasets and OOD generalization tasks.

\begin{acks}
  This work was in part supported by the National Natural Science Foundation of China (62222607), and the Shanghai Municipal Science and Technology Major Project (2021SHZDZX0102).
\end{acks}

	\clearpage

	\balance
	

	\clearpage
	
	\appendix

	\section{Proof of Theorem \ref{th:mixed_feature} and \ref{th:mixed_label}}
	To simplify the analysis, assume that we use the invert of the degree of central node as mixing weight, i.e., $ e_{ij}=1/deg(i) $. Then, we have
	\begin{align*}
		\begin{aligned}
			\mathbb{E}[\mathbf{h}_i]&=\mathbb{E} \left[ \sum_{j\in \mathcal{N}(i) } \frac{1}{deg(i)} \mathbf{h}_j \right] \\
			&=\frac{1}{deg(i)}\sum_{j\in \mathcal{N}(i) } \mathbb{E}[\mathbf{h}_j] \\
			&=\frac{1}{deg(i)}\sum_{j\in \mathcal{N}(i) } \sum_{c\in C} \mathbb{P}(y_j=c|y_i) \boldsymbol{\mu}(c) \\
			&=p \mathbf{\boldsymbol{\mu}}(y_i) + \frac{1-p}{C-1} \sum_{c \neq y_i} \mathbf{\boldsymbol{\mu}}(c).
		\end{aligned}
	\end{align*}
	To prove Eq. \eqref{eq:feature_distance}, we first introduce the Hoeffding's inequality.
	\begin{lemma}[Hoeffding's Inequality] \label{lemma:hoeffding}
		Let $ X_1,\ldots,X_n $ be independent random variables such that $ a\leq X_i \leq b $ for all $ i $. Then
		\begin{align*}
			\mathbb{P}\left( \left| \frac{1}{n} \sum_{i=1}^n (X_i- \mathbb{E}[X_i] ) \right| \geq t \right) \leq 2 \exp\left(-\frac{2nt^2}{(b-a)^2 }  \right)
		\end{align*}
		for all $ t\geq 0 $.
	\end{lemma}
	Let $ \mathbf{h}_i[k] $ denote the $ k $-th element of $ \mathbf{h}_i $. If $ \| \frac{1}{deg(i)} \sum_{j\in \mathcal{N}(i)} (\mathbf{h}_i -\mathbb{E}[\mathbf{h}_i] ) \|_2 \geq \sqrt{F}t_1 $, then at least for one $ k \in \{1,\ldots,F\} $, the inequality $ \left| \frac{1}{deg(i)} \sum_{j\in \mathcal{N}(i)} (\mathbf{h}_i[k] -\mathbb{E}[\mathbf{h}_i[k]] )  \right| \geq t_1 $ holds. Therefore,
	\begin{align*}
		\begin{aligned}
			&\mathbb{P}\left( \left\| \frac{1}{deg(i)} \sum_{j \in \mathcal{N}(i) } (\mathbf{h}_j- \mathbb{E}[\mathbf{h}_j] ) \right\|_2 \geq \sqrt{F}t_1	\right) \\
			\leq& \mathbb{P}\left(\bigcup_{k=1}^F\left\{  \left| \frac{1}{deg(i)} \sum_{j \in \mathcal{N}(i) } (\mathbf{h}_j[k]- \mathbb{E}[\mathbf{h}_j[k]] ) \right| \geq t_1 \right\}  \right) \\
			\leq & \sum_{k=1}^F \mathbb{P}\left( \left| \frac{1}{deg(i)} \sum_{j \in \mathcal{N}(i) } (\mathbf{h}_j[k]- \mathbb{E}[\mathbf{h}_j[k]] ) \right| \geq t_1 \right) \\
			\leq &2F\exp\left(-\frac{deg(i)t_1^2}{2B^2} \right) \quad \quad \quad \quad \text{(By Hoeffding's inequality)}.
		\end{aligned}
	\end{align*}
	Let $ t=\sqrt{F}t_1 $, then we have
	\begin{align*}
		\mathbb{P}\left( \left\| \frac{1}{deg(i)} \sum_{j \in \mathcal{N}(i) } (\mathbf{h}_j- \mathbb{E}[\mathbf{h}_j] ) \right\|_2 \geq t	\right) \leq 2F\exp\left( -\frac{deg(i)t^2}{2B^2F}\right).
	\end{align*}
	The LHS of the above equation is equal to $ \mathbb{P}(\| \mathbf{h}_i - \mathbb{E}[\mathbf{h}_i ]  \|_2 \geq t  ) $, so we complete the proof of Theorem \ref{th:mixed_feature}. 
	
	Next we derive $  \mathbb{E}[\bar{\mathbf{y}}_i ] $ (w.l.o.g, assume $ y_i=c $).
	\begin{align}
		\begin{aligned}
		& \mathbb{E}[\bar{\mathbf{y}}_i ] \\
		=&  \mathbb{E} \left[ \sum_{j\in \mathcal{N}(i) } \frac{1}{deg(i)} \bar{\mathbf{y}}_j \right] \\
		=&\frac{1}{deg(i)}\sum_{j\in \mathcal{N}(i) } \mathbb{E}[\bar{\mathbf{y}}_j] \\
		=&p(1-\epsilon)\mathbf{e}_c+p\frac{\epsilon}{C-1}\sum_{j\neq c}\mathbf{e}_j + \frac{1-p}{C-1}\sum_{j\neq c} \left[(1-\epsilon)\mathbf{e}_j + \frac{\epsilon}{C-1}\sum_{k\neq j} \mathbf{e}_k \right] \\
		=& p(1-\epsilon)\mathbf{e}_c+\frac{p\epsilon+(1-p)(1-\epsilon)}{C-1}\sum_{j\neq c}\mathbf{e}_j + \frac{\epsilon(1-p)}{(C-1)^2}\sum_{j\neq c} \sum_{k\neq j} \mathbf{e}_k.
		\end{aligned} \label{eq:plug1}
	\end{align}
	Note that
	\begin{align}
		\sum_{j\neq c} \sum_{k\neq j} \mathbf{e}_k = (C-2)\sum_{j\neq c} \mathbf{e}_j + (C-1) \mathbf{e}_c.		\label{eq:plug2}
	\end{align}
	Substituting Eq. \eqref{eq:plug2} into Eq. \eqref{eq:plug1} and rearranging the resulting expression, we obtain:
	\begin{align*}
			\mathbb{E}[\bar{\mathbf{y}}_i]	= \left[ p(1-\epsilon)+ \frac{\epsilon(1-p)}{C-1}   \right] \mathbf{e}_c+ \left[\frac{p\epsilon+(1-p)(1-\epsilon)}{C-1} \right. \\
			\left. +\frac{\epsilon(1-p)(C-2)}{(C-1)^2} \right] \sum_{j\neq c} \mathbf{e}_j,
	\end{align*}
	which completes the proof of Eq. \eqref{eq:exp_label}. 
	
	The proof of Eq. \eqref{eq:label_dist} in Theorem \ref{th:mixed_label} closely resembles the proof of Eq. \eqref{eq:feature_distance} in Theorem \ref{th:mixed_feature}, which uses the Hoeffding's inequality. The main distinction lies in the feature dimension being $C$ instead of $F$, and the bound's value being 1 instead of $B$.
	
	\section{Proof of Theorem \ref{th:mixup_gcn}}
	The first-order derivative of $ F_1(\mathbf{H}; \mathbf{H}^{(t)}) $ w.r.t $ \mathbf{h}_u $ is
	\begin{align}
		\frac{\partial F_1(\mathbf{H}; \mathbf{H}^{(t)}) }{\partial \mathbf{h}_u} = 2 (\mathbf{h}_u - \mathbf{h}_u^{(t)} ) + 2 \beta \sum_{(u,v)\in \mathcal{E} } e_{uv} (\mathbf{h}_u-\mathbf{h}_v).		
	\end{align}
	Applying the Forward Euler method with step size $ \tau $ to PDE
	\begin{align}
		\frac{\partial \mathbf{h}_u^{(t)} }{\partial t}=-\frac{\partial }{\partial \mathbf{h}_u}F_1(\mathbf{H}; \mathbf{H}^{(t)}) \label{eq:pde_gcn} ,
	\end{align}
	which updates $ \mathbf{h}_u $ in the direction of gradient descent, we obtain
	\begin{align*}
		\begin{aligned}
			\frac{\mathbf{h}_u^{(t+1)} - \mathbf{h}_u^{(t)}}{\tau} 
			& = - \left. \frac{\partial F_1(\mathbf{H}; \mathbf{H}^{(t)})}{\partial \mathbf{h}_u} \right |_{\mathbf{H} = \mathbf{H}^{(t)}} \\
			& = - 2(\mathbf{h}_u^{(t)} - \mathbf{h}_u^{(t)}) - 2 \beta \sum_{v \in \mathcal{N}(u) } e_{uv} (\mathbf{h}_u^{(t)} - \mathbf{h}_v^{(t)}) \\
			& = - 2\beta \sum_{v \in \mathcal{N}(u)} e_{uv} (\mathbf{h}_u^{(t)} - \mathbf{h}_v^{(t)}).
		\end{aligned}		
	\end{align*}
	After rearranging the equation, we have
	\begin{align*}
		\mathbf{h}_u^{(t+1)} = (1 - 2\tau \beta ) \mathbf{h}_u^{(t)} + 2\tau\beta \sum_{v\in \mathcal{N}(u)} e_{uv} \mathbf{h}_v^{(t)}.		
	\end{align*}
	Setting $ \alpha = 2\tau\beta $ yields  Eq. \eqref{eq:gcn_mixup_cost}.
	
	Substituting $ F_1(\mathbf{H}; \mathbf{H}^{(t)}) $ in Eq. \eqref{eq:pde_gcn} with $ F_2(\mathbf{H}; \mathbf{H}^{(0)}) $, we obtain
	\begin{align*}
		\begin{aligned}
			\frac{\mathbf{h}_u^{(t+1)} - \mathbf{h}_u^{(t)}}{\tau} 
			& = - \left. \frac{\partial F_2(\mathbf{H}; \mathbf{H}^{(0)})}{\partial \mathbf{h}_u} \right |_{\mathbf{H} = \mathbf{H}^{(t)}} \\
			& = - 2(\mathbf{h}_u^{(t)} - \mathbf{h}_u^{(0)}) - 2 \beta \sum_{v \in \mathcal{N}(u) } e_{uv} (\mathbf{h}_u^{(t)} - \mathbf{h}_v^{(t)}).
		\end{aligned}		
	\end{align*}
	Rearranging the equation yields
	\begin{align*}
		\mathbf{h}_u^{(t+1)}=(1-2\tau-2\tau \beta )\mathbf{h}_u^{(t)}+2\tau \mathbf{h}_u^{(0)}+2\tau \beta \sum_{v\in \mathcal{N}(u) } e_{uv} \mathbf{h}_v^{(t)}		.
	\end{align*}
	Setting $ \tau = \frac{1}{2(\beta+1)} $ and $ \alpha = 2\tau $ yields Eq.  \eqref{eq:geomix_app_x}.

	\section{Additional Experimental Details}
	For \texttt{STL10}, we utilize all 13,000 images, each categorized into one of the ten classes. For \texttt{CIFAR10}, we choose 1,500 images from each of 10 classes and obtain a total of 15,000 images. In these two image datasets, we randomly select 10/20 images per class as training set, 4,000 images in total as validation set and the remaining instances as testing set. We also evaluate our model on \texttt{20News}, which is a text classification dataset consisting of 9,607 instances. We follow \citep{franceschi2019learning} to take 10 classes from 20 and use words (TF-IDF) with a frequency of more than 5\% as features. In this dataset, we randomly select 100/200 instances per class as training set, 2,000 instances in total as validation set and the remaining ones as testing set.
	
	We implement our approach using PyTorch. All experiments are conducted on an NVIDIA GeForce RTX 2080 Ti with 11GB memory.
	
	Grid search is used on validation set to tune the hyper-parameters. The learning rate is searched in \{0.001, 0.005, 0.01, 0.05\}; dropout rate is searched in \{0, 0.2, 0.3, 0.5, 0.6\}; weight decay is searched in [1e-5, 1e-2].  Other hyper-parameters for specific models are listed below.
	\begin{itemize}
		\item GCN:  Hidden dimension $ \in $ \{16, 32, 64\}; number of layers is 2.
		\item GAT: Hidden dimension $ \in $ \{8, 16, 32, 64\}; number of heads $ \in $ \{4, 6, 8\}; number of layers is 2.
		\item SGC: Hops $ \in  $ \{2, 3\}.
		\item APPNP: Hidden dimension $ \in $ \{16, 32, 64\}; $ \alpha \in $ \{0.1, 0.2, 0.5\}; hops $ \in $ \{5, 10\}.
		\item GloGNN:  $\alpha \in [0,1]$, $\beta_1 \in \{0,1,10\}$, $\beta_2 \in \{01,1,10,100,1000\}$, $\gamma \in [0,0.9]$, number of norm layers $\in \{1,2,3\}$, $K \in [1,6]$.
		\item Mixup: We set the number of layers to 3, as suggested by its author. The hidden dimension used by the author is 256, and we search it in \{64, 128, 256\}. We search $ \alpha $ in [0.5, 5].
		\item GraphMixup: Hidden dimension $ \in $ \{16, 64, 128\}; number of layers is 2; semantic relation $ K=4 $; loss weights $ \alpha=1.0 $.
		\item GraphMix: We use the same number of layers and hidden size as underlying GCN, as suggested by its author; $ \alpha \in$ \{0.0, 0.1, 1.0, 2.0\}; $ \gamma \in $ [0.1, 10]; temperature $ T=0.1 $; number of permutations $ K=10 $.
		\item EERM: weight for combination $ \beta \in $ \{0.2, 0.5, 1.0, 2.0, 3.0\}; number of edge editing for each node $ s \in $ \{1, 5, 10\}; $ K \in$ \{3, 5\}; number of iterations for inner update $ T \in $ \{1, 5\}; we use the same hidden dimension and number of layers as GCN.
		\item DANN: $ \lambda \in $ [0.2, 6]; number of layers is 2; hidden dimension $ \in $ \{16, 32, 64\}.
		\item \ours-I and \ours-II: Mixup hops (number of consecutive Mixup) $ K \in $ \{1, 2, 3, 4\}; $ \alpha \in $ [0.1, 0.8]; the number of layers and hidden dimension are the same as GCN; $\lambda$ is set to 1 except in a few experiments.
		\item \ours-III: Mixup hops $ K \in $ \{1, 2, 3, 4\}; $ \alpha \in $ [0.1, 0.8]; graph weight $ \eta \in $ \{0.3, 0.5, 0.7, 0.8\}; projection dimension $ F'=16 $; the number of layers and hidden dimension are the same as GCN; $\lambda$ is set to 1.
	\end{itemize}
	
	\section{Hyper-parameter Analysis}
\begin{figure}[tb!]
		\centering
		\includegraphics[width=\linewidth]{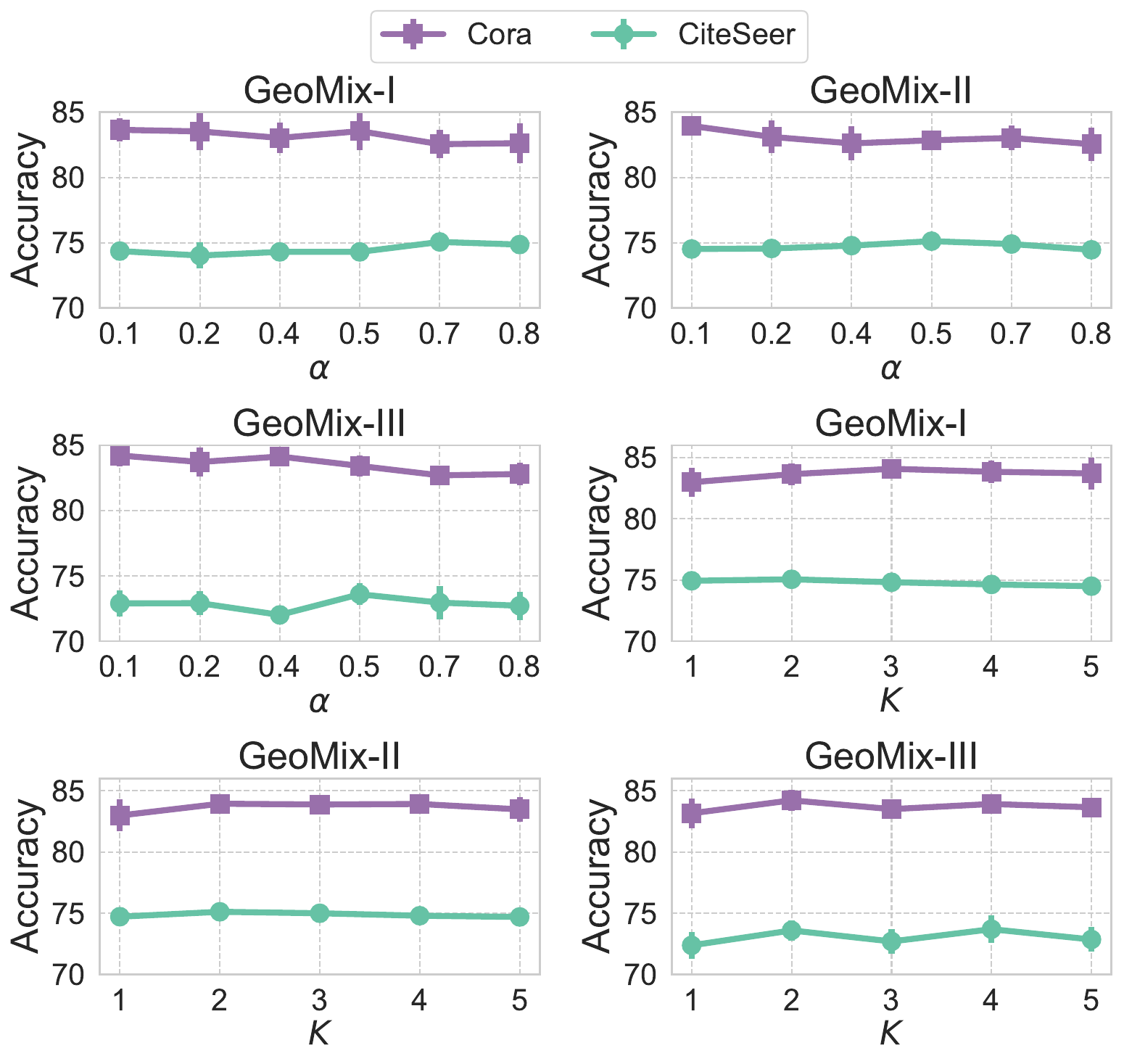}
		\caption{Impacts of $ \alpha $ and Mixup hop $ H $ on Geometric Mixup.}
		\label{fig:hyper}
	\end{figure}

	We present the accuracy of Geometric Mixup concerning both $ \alpha $ and Mixup hops $ K $ in Fig. \ref{fig:hyper}. Remarkably, using 2 hops generally yields the most competitive performance. Therefore, by setting Mixup hops to 2, we can achieve desirable performance with minimal overhead.


\end{document}